# High-Level Text Preprocessing for Semantic Similarity Analysis of Discursive Texts: A Framework and Empirical Demonstration

**Mehmet Murat Albayrakoglu, PhD**
Adjunct Instructor
Işık University
Faculty of Economics, Administrative and Social Sciences
Department of Management Information Systems
Istanbul, Türkiye
ORCID: 0000-0002-5057-5641

**Mehmet Nafiz Aydin, PhD**
Professor, Head of Department
Boğaziçi University
Faculty of Managerial Sciences,
Department of Management Information Systems
Istanbul, Türkiye
ORCID: 0000-0002-3995-6566

## Abstract

Semantic Textual Similarity (STS) methods assume that a document's lexical content faithfully represents what it asserts. This assumption fails for discursive documents that discuss, compare, critique, and contextualize other positions in the process of articulating their own. The result is *semantic diffusion*: similarity scores between documents are inflated by vocabulary acquired through discursive engagement rather than substantive alignment. Standard Natural Language Processing (NLP) preprocessing (tokenization, stopword removal, stemming, lemmatization) cannot address this problem because it operates at the lexical level, treating all content identically regardless of its discursive function. This paper introduces *high-level text preprocessing*: a systematic, rule-based intervention applied before the standard preprocessing pipeline to isolate each document's actual claim from its discursive structure. We propose 12 rules, each with an explicit rationale, and demonstrate their effect on an encyclopedic philosophical corpus: three entries from the Stanford Encyclopedia of Philosophy (virtue ethics, deontological ethics, and consequentialism). A three-phase experiment using eight Transformer-based STS models shows that preprocessing reduces centroid cosine similarity scores across all three theory pairs, with 23 of 24 model-pair comparisons showing the expected decrease and cross-model agreement ranging from 7-1 to 8-0. We introduce the *semantic diffusion index* (*SDI*), a per-document metric for assessing the semantic reorientation between a document's raw and high-level preprocessed representations. Although the framework is demonstrated using philosophical texts, it potentially addresses a domain-agnostic problem applicable to legal texts, policy documents, academic articles, and any genre in which a discursive approach introduces vocabulary from positions the document does not endorse.

## 1. Introduction

Whether traditional set-based measures like Jaccard similarity or modern embedding-based approaches using Transformer models, Semantic Textual Similarity (STS) (Agirre et al., 2012) methods rely on a common assumption: the lexical content of a document faithfully reflects what it asserts. The standard preprocessing pipeline—involving tokenization, stopword removal, lemmatization, and stemming—operates mechanically on the text it receives, treating every word in the document as a legitimate contributor to the document's semantic profile.

This assumption holds for many text types. A product review, a news article, or a social media post generally says what it means: the words in the document are the message. However, a large and important class of texts, the so-called *discursive texts*, that do not merely state a position but also discuss, compare, critique, and contextualize other positions in the process of articulating their own, systematically violate this assumption.

In this regard, philosophical texts are exemplary. An encyclopedia entry on virtue ethics does not merely describe it. It explains how virtue ethics differs from deontological ethics and consequentialism. It discusses criticisms from other frameworks and responds to them. It names some other philosophers who provided alternative views. As a result, the raw text of the virtue ethics entry contains substantial vocabulary drawn from deontological ethics and consequentialism, not because virtue ethics conflates those frameworks, but because the discursive nature of philosophical writing requires engaging with them.

Legal texts exhibit similar properties. A statute's preamble often references the policy landscape it aims to reform, the legislative precedents it supersedes, and the competing interests it seeks to balance. Academic articles in any field routinely dedicate significant sections to positioning their contribution against prior work by incorporating additional vocabulary from the very approaches they seek to distinguish themselves from.

When such texts are submitted to an STS pipeline without prior intervention, the result is *semantic diffusion*: the similarity scores between documents are inflated by vocabulary that reflects discursive engagement rather than substantive alignment. A virtue ethics text and a deontological ethics text will appear more similar than they substantively are, because each contains vocabulary acquired from the other's domain through argumentation, not through genuine conceptual overlap.

Standard NLP preprocessing cannot address this problem. Tokenization does not ask whether a sentence belongs in the document for the purposes of the analysis. Stopword removal does not distinguish between a word used assertively and a word used contrastively. Lemmatization collapses inflected forms but cannot tell whether a lemma represents the document's own position or an opposing position. The entire standard pipeline assumes that content has already been selected and that the text, as given, is the right text to analyze.

This paper introduces high-level text preprocessing: a systematic, rule-based intervention applied to discursive texts before the standard NLP preprocessing pipeline to isolate the assertive core of each document—what it actually claims—from its discursive elements—what it discusses, compares against, argues about, or merely references. The rules are designed to minimize semantic diffusion across documents that will subsequently be compared using one or more of the set-theoretic, vector-based, or embedding-based STS methods.

The NLP literature on text preprocessing is extensive (Chai, 2023; Jurafsky & Martin, 2025). Standard preprocessing pipelines are well understood and widely implemented. Their purpose is to normalize a text's lexicon by eliminating variant forms of the same word, thereby reducing orthographical and morphological variation, so that downstream models can focus on semantic content. What this literature does not address is a logically prior question: whether the text being preprocessed is the *right* text to preprocess.

However, in discursive texts, content and assertion diverge. For example, a philosophical encyclopedia entry about a theory of ethics contains three distinct types of material:

1. **Assertive content—**statements that describe what the theory is, its principles, its commitments, and its characteristic claims.
2. **Contrastive content—**statements that describe what the theory is not, or how it differs from other theories, such as comparisons, objections, and responses.
3. **Meta-discursive content—**statements about the document itself—titles, abstracts, tables of contents, reference lists, editorial apparatus.

Only the first category, assertive content, represents the theory's substantive semantic profile. The second introduces vocabulary from other theories, creating cross-document semantic diffusion. The third adds noise unrelated to any theory. Standard preprocessing pipelines treat all three categories equally. They cannot distinguish a sentence that asserts "virtue ethics emphasizes character" from a contrastive sentence that says "unlike deontological ethics, which emphasizes duty," or from a meta-discursive statement, such as "see Section 3 for further discussion."

In prior work, applying semantic similarity methods to the relationship between normative theories of ethics and organizational codes of conduct (Albayrakoglu & Aydin, 2022), this problem was addressed by removing references to other theories and examples from the source texts before computing similarity. That study described preprocessing as comprising linguistic processing and contextual processing (Albayrakoglu & Aydin, 2022; cf.

Rozeva & Zerkova, 2017), but did not formalize the latter. The present paper develops the contextual dimension into an explicit, rule-based framework and demonstrates its effect using Transformer-based models.

This paper aims to make the following contributions:

1. **Identification of the gap.** We identify a systematic gap in the text preprocessing literature: a lack of methods for isolating assertive content from discursive content prior to similarity analysis.
2. **A rule-based preprocessing framework.** We propose twelve high-level preprocessing rules, each with an explicit rationale, designed to minimize semantic diffusion across documents that will subsequently be compared using STS methods.
3. **Empirical demonstration.** We seek to demonstrate the effect of high-level preprocessing on similarity scores by comparing raw and preprocessed versions of three normative theories of ethics drawn from the Stanford Encyclopedia of Philosophy, using eight Transformer-based STS models.
4. **The semantic diffusion index.** We introduce the SDI, a per-document metric, indicating the semantic reorientation between a document's raw and high-level preprocessed representations
5. **Domain-agnostic motivation, domain-specific demonstration.** While the rules will be illustrated through philosophical texts, the underlying problem applies to any domain where texts discuss, compare, or argue against other positions.

## 2. Related Work

Text preprocessing is a well-established stage in NLP pipelines, and its impact on downstream tasks has been extensively studied. The standard pipeline comprises tokenization, lowercasing, stopword removal, stemming, and lemmatization (Chai, 2023; Jurafsky & Martin, 2025). Uysal and Gunal (2014) systematically examined all possible combinations of standard preprocessing methods on four datasets across two languages and demonstrated that the choice of preprocessing combination significantly affects classification accuracy. Subsequent comparative work has reinforced this picture (Balakrishnan & Lloyd-Yemoh, 2014; Pramana et al., 2022; Chai, 2023). In all cases, the methods under study are lexical-level transformations.

Alshanik et al. (2020) proposed a hyperplane-based method for automatically extracting domain-specific common words that lack discriminative information within a given corpus. This represents a step toward content-aware preprocessing, but it still operates at the word level, as the method identifies uninformative tokens rather than uninformative passages. The critical observation is that preprocessing is conceived as a lexical-level operation. No method in the standard preprocessing literature asks whether a given sentence, paragraph, or passage should be included in the document for the purposes of a particular analysis.

A separate body of work, argument mining, operates at the content level. It aims to automatically identify and extract the structure of inference and reasoning in natural language (Lippi & Torroni, 2016; Lawrence & Reed, 2019). More recent work has extended these capabilities using Transformer-based models (Cheng et al., 2022; Irani et al., 2025). In principle, these methods distinguish between what a document asserts and what it discusses. However, this potential has not been realized for document curation prior to similarity analysis. No study in the argument mining literature applies the output of claim detection or stance classification as a preprocessing step for STS.

Within the STS literature, preprocessing is treated as a brief methodological footnote. Studies typically report that texts were "tokenized and lemmatized" without further elaboration (Chandrasekaran & Mago, 2021; He et al., 2024; Wang & Dong, 2020). This assumption is rarely questioned because the dominant text types in STS research are not discursive.

The gap lies at the intersection of these three bodies of work:

1. Standard preprocessing operates at the lexical level and cannot distinguish assertive content from contrastive or meta-discursive content.
2. Argument mining has developed content-level tools but has never applied them to document curation for similarity analysis.
3. STS studies assume that the input text faithfully represents the document's assertions, an assumption that fails for discursive texts.

This paper addresses the gap by proposing high-level text preprocessing: a systematic, content-level intervention applied before the standard lexical preprocessing pipeline.

## 3. The Problem: Discursive Texts and Semantic Diffusion

Discursive texts contain three functionally distinct types of material, as presented in the Introduction:

1. Assertive content comprises statements that articulate what the document's subject is, claims, or entails.
2. Contrastive content comprises statements that articulate what the document's subject is not, or how it differs from other positions. These introduce vocabulary from other domains, creating cross-document semantic diffusion.
3. Meta-discursive content comprises statements about the document itself, its structure, organization, and bibliographic apparatus.

Standard NLP preprocessing operates at the lexical level rather than the content level. Therefore, it cannot identify the discursive function of a sentence based on its substantive-versus-comparative/contrastive role and decide whether to include or exclude it. The problem requires a content-level intervention that precedes the standard pipeline. This is what we call *high-level text preprocessing*.

Philosophical texts exacerbate the problem by combining linguistic variability with extensive discursive scaffolding. Standard preprocessing techniques, stemming and lemmatization, address some of this variability at the lexical level. What they cannot address is the structural problem identified in this paper: discursive texts contain entire passages whose vocabulary belongs to positions the document does not endorse. The high-level preprocessing primarily focuses on discursive scaffolding. Encyclopedic philosophical texts are an ideal test case because they strongly exhibit the problem. We use three entries from the Stanford Encyclopedia of Philosophy (SEP): virtue ethics (Hursthouse & Pettigrove, 2023), deontological ethics (Alexander & Moore, 2021), and consequentialism (Sinnott-Armstrong, 2023).

## 4. High-Level Preprocessing Rules

The following twelve rules were developed iteratively during the preprocessing of three encyclopedic philosophical texts. Each rule targets a specific source of semantic diffusion or noise. The rules are divided into four categories: The first two categories target non-assertive content and aim to eliminate them: meta-discursive (three rules) and contrastive (four rules). The next category involves lexical normalization (four rules). Lexical normalization standardizes how concepts are represented at the word level across documents, so that the same idea is expressed using the same tokens regardless of which document contains it. It eliminates surface-level differences, such as those in spelling, language, or attribution. Thus, it prevents incorrect lexical overlap or divergence unrelated to theoretical content. The last category is assertive compression (one rule), intended to condense each argument to its conclusion.

**Rule 1: Remove titles, subtitles, and section headers.**

*Rationale.* They indicate the structure of the document, not the content of the theory.

**Rule 2: Eliminate meta-descriptions.**

*Rationale.* Descriptions of the document describe what it is about or how it is organized, not what the theory claims.

**Rule 3: Remove items from the reference list and citations in the text.**

*Rationale.* Bibliographic entries and citations include author names, publication titles, and publication years, which add noise.

**Rule 4: Delete descriptions of, references to, or comparisons with other theories.**

*Rationale.* Rule 4 is the broadest and most consequential rule. It targets any passage that introduces vocabulary from a theory other than the one being profiled. Explicit comparisons, implicit contrasts, and historical narratives all introduce vocabulary from rival theories.

The concern that Rule 4 "fundamentally alters the source texts" is correct and intentional. High-level preprocessing is *designed* to alter the source texts by separating their assertive core from their discursive scaffolding. If the purpose is to measure substantive similarity—what each theory claims—then passages in which one theory discusses another are evidence of what it *engages with*, not what it *claims*. Retaining them conflates substantive alignment with discursive engagement.

Rule 4 works in conjunction with Rule 7: passages comparing forms of the same theory are retained, while cross-theory content is removed.

**Rule 5: Delete discussions about what the theory is not, but keep negative examples.**

*Rationale.* Discussions of what a theory is not constitute the primary vehicle for semantic diffusion. Transformer-based STS models are known to have limited sensitivity to negation (Hossain et al., 2020; Ettinger, 2020). A sentence asserting "virtue ethics emphasizes rules" and a sentence asserting "virtue ethics does not emphasize rules" produce similar embeddings because they share the same content words. Leaving contrastive passages in the text allows vocabulary from rival theories to diffuse the semantic profile in exactly the way the models are least equipped to correct.

The critical qualification—*but keep negative examples*—distinguishes negation-as-contrast (what the theory is not, relative to other theories) from negation-as-assertion (what the theory says people should not do). Only the former introduces cross-theory vocabulary.

The objection that these theories "define themselves" by contrast is accurate as a description of philosophical method, but does not entail that contrastive content should be retained for STS. The question is not whether contrastive passages are philosophically important, but whether retaining them would produce similarity scores that reflect what each theory *asserts* rather than what it *engages with*.

**Rule 6: Remove references to religions and religious symbols.**

*Rationale.* Ensure religious neutrality. Historical connections to religious traditions are not part of the theories' philosophical content for secular semantic comparison.

**Rule 7: Keep comparisons of the various forms of the same theory.**

*Rationale.* Internal comparisons describe the theory's own landscape and are indispensable extensions of its semantic profile.

**Rule 8: Remove proper nouns.**

*Rationale.* The purpose of removing proper nouns is to eliminate a class of tokens whose distribution across documents reflects attribution patterns rather than theoretical content. In discursive philosophical texts, proper nouns appear primarily in attributive contexts: "Kant argued that," "Aristotle held that," "Mill proposed." Each theory's entry names its own proponents and its interlocutors from rival traditions, and the distribution of these names varies for reasons unrelated to theoretical substance.

The virtue ethics entry mentions Aristotle (20 times), Hume (14), Anscombe (6), and Foot (8) from its own tradition, as well as Kant (4) and Mill (9) from rival traditions in contrastive passages. The deontological ethics entry mentions Kant (18), Nagel (8), Scanlon (5), and Rawls (4) from its own tradition, but also Anscombe (6) and Foot (4) from virtue ethics and Bentham (2) from consequentialism. The consequentialism entry mentions Mill (12), Bentham (8), Sidgwick (8), Singer (7), and Railton (5) from its own tradition, but also Rawls (4), Ross (3), and Foot (2) from rival traditions. The shared proper nouns create lexical overlap that Transformer models encode as semantic similarity, but the overlap reflects whom the entries talk about, not what they claim. Retaining proper nouns would also introduce a confound: theories whose proponents overlap with those of other traditions would generate more shared tokens for reasons independent of theoretical content.

**Rule 9: Convert text into US English.**

*Rationale.* Eliminate spelling variations that create spurious lexical differences.

**Rule 10: Replace foreign-language words with their US-English equivalents, if any.**

*Rationale.* Replacing foreign terms reduces vocabulary gaps in the model.

**Rule 11: Add brief English definitions of foreign-language words if there is no US-English equivalent.**

*Rationale.* Adding a definition ensures the concept contributes to the lexical profile.

**Rule 12: Keep only the conclusive statements for incremental arguments.**

*Rationale.* Only the conclusion captures the theory's actual claim. Retaining intermediate steps inflates the vocabulary with transitional terms.

Table 1 summarizes the 12 rules discussed above.

**Table 1** Summary of the high-level preprocessing rules

| Category | Rule | Definition | Purpose |
|---|---|---|---|
| *Meta-discursive* | 1 | Remove titles, subtitles, and section headers. | Remove content about the document itself, such as its structure, organization, and bibliographic apparatus. |
| | 2 | Eliminate meta-descriptions. | |
| | 3 | Remove reference list items and in-text citations. | |
| *Contrastive* | 4 | Delete cross-theory descriptions, references, and comparisons. | Remove vocabulary from rival theories introduced through discursive engagement. Exception: Rule 7 is a retention rule that qualifies Rule 4. |
| | 5 | Delete what the theory is not; keep negative examples. | |
| | 6 | Remove religious references | |
| | 7 | Retain internal comparisons (exception to Rule 4). | |
| *Normalizing* | 8 | Remove proper nouns. | Reduce spurious lexical variation across documents unrelated to theoretical content. |
| | 9 | Convert to US English. | |
| | 10 | Replace foreign words with equivalents. | |
| | 11 | Add definitions of untranslatable foreign words. | |
| *Condensing* | 12 | Keep only conclusive statements. | Condense multi-step arguments to their conclusions, removing intermediate scaffolding vocabulary. |

## 5. Corpus and Framework Application

### 5.1. Corpus Selection

The raw corpus consists of three entries from the Stanford Encyclopedia of Philosophy (SEP): Virtue Ethics (Hursthouse & Pettigrove, 2023), Deontological Ethics (Alexander & Moore, 2021), and Consequentialism (Sinnott-Armstrong, 2023). A critical requirement is that each theory be represented by a consolidated, self-contained treatment of comparable length. SEP entries satisfy this: each exceeds 12,000 words (range: 12,801–13,259) within a single document. Alternative sources—including the Routledge Encyclopedia of Philosophy (Craig, 1998), the Encyclopedia of Ethics (Becker & Becker, 2001), Encyclopaedia Britannica, the Internet Encyclopedia of Philosophy, the Oxford Handbook of Ethical Theory (Copp, 2006), and the Bloomsbury Handbook of Ethics (Miller, 2023)—were evaluated but found unsuitable. A detailed assessment of the alternative corpora is provided in Appendix A.

The use of three texts from a single domain is a deliberate design choice: the paper proposes a framework and demonstrates its effect, rather than claiming cross-domain generalizability. Three mutually discursive texts constitute the minimum configuration for detecting differential preprocessing effects across pairs and computing independent per-document SDIs. At the same time, eight Transformer models serve as the basis for assessing consistency.

Encyclopedic philosophical texts are chosen for several reasons. First, a single idea can be articulated through diverse linguistic formulations, in philosophical discourse (Rohatyn, 1972), introducing inherent ambiguities that lead to varying interpretations and complicate computational approaches to meaning (Adler & Van Doren, 1972; Gray, 2012; Martinich, 2016). Second, the longer the text, the greater the variability and the greater the analytical challenge (Jurafsky & Martin, 2025). Third, their discursive structure may include entire passages whose vocabulary belongs to positions the document does not endorse. Standard preprocessing techniques address some linguistic variability at the lexical level but cannot resolve this structural problem. High-level preprocessing targets the latter challenge; therefore, encyclopedic philosophical texts provide an ideal test case.

### 5.2. Application of the High-Level Preprocessing Rules

SEP entries are demonstrably discursive—preprocessing removed 47-71% of the raw content across the three entries, making them a suitable test case for the framework. Whether they represent a high, typical, or low level of discursiveness relative to other genres is itself an empirical question that the SDI could answer in future cross-domain work. Encyclopedic philosophical texts are, on the one hand, an ideal test case because they clearly exhibit the problem, and, on the other hand, they share structural and stylistic conventions that reduce extraneous variation across documents obtained from different sources.

The twelve rules were not applied as a pre-existing checklist. They emerged iteratively during the manual preprocessing: as each entry was processed, recurring patterns of non-assertive content were identified, articulated as rules, and then applied backward to entries already partially processed. Because individual rules were not applied in isolation and their effects were not recorded separately, ablation statistics per rule are not available. This is a limitation of the iterative method (discussed in Section 8.4). It is also a feature: the rules emerged from the material rather than being imposed on it. Table 2 presents the corpus characteristics for all three entries before and after preprocessing. On the table, CO denotes consequentialism, DE denotes deontological ethics, and VE denotes virtue ethics.

**Table 2** Corpus characteristics before and after high-level preprocessing.

| | Words | | | Sentences | | |
|---|---|---|---|---|---|---|
| **Theory** | **Raw No.** | **Remaining No.** | **Reduction %** | **Raw No.** | **Remaining No.** | **Reduction** |
| *CO* | 12,953 | 6,892 | 46.8% | 812 | 278 | 65.8% |
| *DE* | 12,801 | 4,583 | 64.2% | 544 | 173 | 68.2% |
| *VE* | 13,259 | 3,828 | 71.1% | 677 | 147 | 78.3% |
| **Total** | **39,013** | **15,303** | **60.8%** | **2,033** | **598** | **70.6%** |

The three raw entries are nearly identical in length (12,801-13,259 words). After preprocessing, they diverge substantially: virtue ethics shows a 71.1% reduction in word count, deontological ethics shows a 64.2% reduction, and consequentialism shows a 46.8% reduction. The ranking of word reduction percentages (VE > DE > CO) reflects the discursive structure of the entries. The percentage reductions in the number of sentences are higher than those of words for all three theories: 78.3% for virtue ethics, followed by 68.2% for deontological ethics, and 65.8% for consequentialism. The ranking of sentence reduction percentages (VE > DE > CO) remains the same as the word reduction percentages, pointing to the discursive nature of the entities.

### 5.3. A Posteriori Transformation Audit

Although iterative development and application of the rules did not permit ablation statistics, a retrospective, order-preserving lexical alignment of each raw-preprocessed text pair was performed to characterize the transformation. The results of the a posteriori transformation audit, which characterizes the ensuing changes, are shown in Table 3. The table compares each raw-preprocessed pair using order-preserving lexical alignment.

Across the three preprocessed texts, approximately 98.24% of lexical tokens were carried over unchanged from the corresponding raw texts; only approximately 1.76% appeared as new or replacement tokens. The latter category includes all lexical normalization, definitional additions, grammatical repairs, proper-name substitutions, and condensation edits. Thus, Rules 9–11 constitute only a subset of this already limited transformation. This analysis does not provide per-rule ablation, but establishes that high-level preprocessing was overwhelmingly subtractive rather than additive or rewriting-based.

**Table 3** A posteriori transformation audit.

| Theory | Preprocessed lexical tokens | Retained unchanged | Retained unchanged (%) | New and replacement | New and replacement (%) |
|---|---|---|---|---|---|
| *VE* | 3,859 | 3,794 | 98.32% | 65 | 1.68% |
| *DE* | 4,601 | 4,501 | 97.83% | 100 | 2.17% |
| *CO* | 6,879 | 6,774 | 98.47% | 105 | 1.53% |
| ***Corpus*** | **15,339** | **15,069** | **98.24%** | **270** | **1.76%** |

## 6. Experiment Design

The experiment tests whether high-level text preprocessing systematically changes the pairwise semantic similarity scores between discursive texts. The corpus consists of two versions of each encyclopedic entry: the raw originals and preprocessed versions resulting from the manual application of the twelve rules. Both versions undergo the same sentence segmentation and model-specific encoding procedure before comparison.

### 6.1 Transformer models

The experiment uses eight Transformer-based STS models, spanning different training objectives and model sizes. Five models are selected for their established use in sentence-level STS tasks: SBERT (Reimers & Gurevych, 2019; Song et al., 2020), Paraphrase-ALBERT (Lan et al., 2020), DistilBERT (Sanh et al., 2019), RoBERTa (Liu et al., 2019), and TinyBERT (Jiao et al., 2020). Three additional models broaden architectural diversity: MiniLM (Wang et al., 2020), E5-base (Wang et al., 2024), and GTE-base (Li et al., 2023). Table 4 summarizes the models' architectures and training characteristics. Sentences exceeding maximum input length in tokens (*Max. seq.*) are truncated, though typical philosophical sentences (25–40 tokens) fall well within all limits. For E5-base, each sentence was prefixed with `query:` in accordance with the model's recommended usage for symmetric semantic-similarity tasks.

### 6.2 Document encoding and similarity metric

Each document is split into sentences using NLTK's `sent_tokenize` function. Empty segments and segments shorter than five characters were discarded. Each sentence was encoded independently; sentences exceeding a model's maximum sequence length remained subject to that model's truncation limit.

Document-level similarity is computed as *centroid cosine similarity*. All sentence embeddings for a document are averaged into a single centroid vector representing the document's overall semantic orientation, and the cosine similarity between two document centroids is taken as the similarity score. A formal mathematical treatment is provided in Appendix B.

**Table 4** Transformer models used in the experiment.
*Parameters* = total learned weights.
*Emb. dim.* = dimensionality of the sentence embedding vector (the $d$ in Appendix B).
*Max. seq.* = maximum input length in tokens.

| Short name | Model identifier | Training objective | Parameters | Emb. dim. | Max seq |
|---|---|---|---|---|---|
| *SBERT* | all-MPNet-base-v2 | Contrastive learning | ~110M | 768 | 384 |
| *ALBERT* | paraphrase-albert-small-v2 | Paraphrase mining | ~12M | 768 | 100 |
| *DistilBERT* | distilbert-base-nli-stsb-mean-tokens | NLI + STS fine-tuning | ~66M | 768 | 128 |
| *RoBERTa* | all-distilroberta-v1 | Contrastive learning | ~82M | 768 | 512 |
| *TinyBERT* | paraphrase-TinyBERT-L6-v2 | Paraphrase mining | ~67M | 768 | 128 |
| *MiniLM* | all-MiniLM-L12-v2 | Knowledge distillation | ~33M | 384 | 128 |
| *E5-base* | intfloat/e5-base-v2 | Contrastive pre-training | ~110M | 768 | 512 |
| *GTE-base* | thenlper/gte-base | Multi-stage contrastive | ~110M | 768 | 512 |

An alternative aggregation strategy, mean pairwise sentence similarity—which averages the cosine similarity across all $m \times n$ sentence pairs, where $m$ and $n$ are the number of sentences in each document—is not used in this study. When two documents each contain hundreds of sentences addressing diverse subtopics, the vast majority of cross-document sentence pairs are semantically unrelated, producing a low noise floor that dominates the mean. Changes in document composition—including the removal of any sentences, regardless of their semantic content—alter the noise floor and shift the mean in ways unrelated to the signal of interest. Such a *concentration effect*, in turn, makes the mean pairwise similarity unsuitable for measuring the semantic impact of content-level preprocessing on long discursive documents. A formal derivation of the concentration effect, and an explanation of why centroid similarity avoids the pair-count concentration mechanism.

### 6.3 Experimental phases and analysis

The experiment is organized into three phases to assess semantic changes before and after preprocessing and quantify the level of change. The phases are as follows:

**Phase 1 (Raw-vs-Raw between theory pairs)** involves three raw entries. Entries are semantically compared pairwise. It establishes the baseline and shows how similar the theories appear when their raw texts, including all contrastive and meta-discursive content, are compared. These scores reflect both genuine conceptual overlap and semantic diffusion, but cannot distinguish between the two. A total of 3 pairs × 8 models = 24 STS scores are calculated.

**Phase 2 (Preprocessed-vs-Preprocessed between theory pairs)** compares the three preprocessed entries pairwise. It measures the same inter-theory similarity after preprocessing has removed non-assertive content. The difference between Phase 1 and Phase 2 quantifies the change in inter-theory similarity associated with high-level preprocessing. Under the semantic-diffusion hypothesis, a positive reduction is consistent with removed discursive and other non-target material having contributed to the raw similarity. This is the primary test that requires another total of 3 pairs of 8 models = 24 STS scores.

**Phase 3 (Raw-vs-Preprocessed per theory)** compares each theory's raw entry with its own preprocessed version. It shifts from inter-document comparison to intra-document diagnosis. It answers the question of how much each document's semantic orientation changes when its non-assertive content is removed. This produces the SDI — a per-document measure independent of what the document is being compared against. The last total of 3 pairs × 8 models = 24 STS scores are calculated.

Thus, the experiment comprises a total of 72 centroid similarity scores. Table 5 summarizes these three phases. The analysis examines four dimensions: score reduction (Phase 1 vs. 2), cross-model consistency, differential effect across pairs, and the semantic diffusion index (Phase 3). The gap between each theory's Phase 3 score and a perfect score of 1.0 defines the semantic diffusion index (SDI), which is the additive complement of the centroid similarity between a document's raw and preprocessed versions.

$$SDI = 1 - \text{raw-vs-preprocessed centroid similarity}$$

A higher SDI indicates a larger semantic reorientation between the raw and preprocessed representations. SDI magnitudes can be compared with pairwise score reductions as complementary indicators, but they do not, by themselves, determine those reductions.

**Table 5** Summary of experimental phases

| Phase | Comparison | Pairs | Question | Expected |
|---|---|---|---|---|
| 1 | Raw-vs-Raw | *VE–DE, VE–CO, DE–CO* | Inter-theory with diffusion | Higher scores |
| 2 | Preprocessed-vs-Preprocessed | *VE–DE, VE–CO, DE–CO* | Inter-theory without diffusion | Lower scores |
| 3 | Raw-vs-Preprocessed | *VE, DE, CO* | Intra-theory (SDI) | High, imperfect |

## 7. Results

### 7.1 Phase 1 versus Phase 2: Inter-theory similarity

Table 6 presents the centroid cosine similarity scores for all three theory pairs under both conditions across all eight models. All three pairs of theories show lower average centroid similarity after preprocessing. VE-DE and VE-CO each achieve a unanimous 8-0 agreement; *DE-CO* achieves 7-1 with only DistilBERT dissenting (by –0.006). Out of 24 comparisons, 23 show the expected decrease.

The magnitude varies: VE-DE and VE-CO show nearly identical average reductions (Δ = 0.039, approximately 4.5% by magnitude), while DE-CO is smaller (Δ = 0.022, 2.4% by magnitude). E5-base and GTE-base produce near-ceiling scores (0.97–0.99), leaving limited room for preprocessing to reduce the score, yet the effect is consistently present (between 0.006 and 0.010).

Overall, five models (ALBERT, RoBERTa, TinyBERT, MiniLM, E5-base) show consistent decreases across all three pairs. GTE-base shows the expected direction but with minimal magnitudes, consistent with a ceiling effect. DistilBERT is the only model showing an increase for any pair: DE-CO increases slightly, and this increase is confined to the pair with the smallest overall effect.

### 7.2 Phase 3: Intra-theory semantic similarity and Semantic Diffusion Indices (SDIs)

Table 7 presents the Phase 3 centroid results. The average SDI ranking is as follows: CO (0.101) > VE (0.094) > DE (0.044), with consequentialism having the highest average SDI and deontological ethics the lowest. However, individual models do not uniformly follow the average ranking: DistilBERT, E5-base, and GTE-base rank VE above CO. What is consistently observed is that DE has the lowest SDI across all eight models.

**Table 6** Centroid cosine similarity: Phase 1 (raw) vs. Phase 2 (preprocessed).
Δ = Phase 1 – Phase 2; positive values indicate a decrease.

| Pair | Model | Phase 1 (Raw) | Phase 2 (Preprocessed) | Difference (Δ) | Direction | Voting |
|---|---|---|---|---|---|---|
| *VE-DE* | SBERT | 0.8147 | 0.7860 | +0.0287 | ↓ | |
| | ALBERT | 0.8538 | 0.7726 | +0.0812 | ↓ | |
| | DistilBERT | 0.8768 | 0.8440 | +0.0328 | ↓ | |
| | RoBERTa | 0.8247 | 0.7704 | +0.0543 | ↓ | |
| | TinyBERT | 0.8354 | 0.7835 | +0.0519 | ↓ | |
| | MiniLM | 0.7691 | 0.7194 | +0.0497 | ↓ | |
| | E5-base | 0.9826 | 0.9741 | +0.0085 | ↓ | |
| | GTE-base | 0.9827 | 0.9766 | +0.0061 | ↓ | |
| | **Average** | **0.8675** | **0.8283** | **+0.0392** | ↓ | **8 against 0** |
| *VE-CO* | SBERT | 0.8221 | 0.7635 | +0.0586 | ↓ | |
| | ALBERT | 0.8744 | 0.8072 | +0.0672 | ↓ | |
| | DistilBERT | 0.9024 | 0.8892 | +0.0132 | ↓ | |
| | RoBERTa | 0.8320 | 0.7630 | +0.0690 | ↓ | |
| | TinyBERT | 0.8741 | 0.8352 | +0.0389 | ↓ | |
| | MiniLM | 0.8332 | 0.7844 | +0.0488 | ↓ | |
| | E5-base | 0.9850 | 0.9754 | +0.0096 | ↓ | |
| | GTE-base | 0.9854 | 0.9764 | +0.0090 | ↓ | |
| | **Average** | **0.8886** | **0.8493** | **+0.0393** | ↓ | **8 against 0** |
| *DE-CO* | SBERT | 0.9086 | 0.8590 | +0.0496 | ↓ | |
| | ALBERT | 0.9151 | 0.8955 | +0.0196 | ↓ | |
| | DistilBERT | 0.9293 | 0.9352 | –0.0059 | ↑ | |
| | RoBERTa | 0.9034 | 0.8715 | +0.0319 | ↓ | |
| | TinyBERT | 0.9011 | 0.8743 | +0.0268 | ↓ | |
| | MiniLM | 0.8873 | 0.8471 | +0.0402 | ↓ | |
| | E5-base | 0.9898 | 0.9827 | +0.0071 | ↓ | |
| | GTE-base | 0.9920 | 0.9862 | +0.0058 | ↓ | |
| | **Average** | **0.9283** | **0.9064** | **+0.0219** | ↓ | **7 against 1** |

The SDI results are broadly consistent with the pairwise reductions but should not be expected to directly predict their magnitudes. The DE–CO pair shows the smallest centroid reduction (0.022), whereas VE–DE and VE–CO show nearly identical larger reductions (0.039). This suggests that pairwise change depends not only on the magnitude of each document's semantic reorientation, as measured by the SDI, but also on the direction of that reorientation relative to the other document.

## 8. Discussion

### 8.1 The preprocessing effect

The consistent reduction across models supports the semantic-diffusion hypothesis. The retrospective text-alignment audit further shows that the transformation was overwhelmingly subtractive, while lexical normalization and additive modification were marginal. The observed score changes can therefore be interpreted primarily as consequences of content-level removal, although the design does not isolate the effect of individual removal rules.

Average reductions range from 2.4% (DE-CO) to 4.5% (VE-DE and VE-CO). These percentages operate on a high baseline (0.77–0.99), reflecting shared normative-ethics vocabulary. The preprocessing effect measures the marginal contribution of discursive engagement to that baseline. The near-equality of VE-DE and VE-CO reductions (0.039 each) reflects VE's role as a discursive hub—the theory that devotes the most space to distinguishing itself from both alternatives. DE-CO's smaller reduction (0.022) reflects less extensive mutual engagement.

### 8.2 Model behavior

The consistency across eight models with different architectures, training objectives, and model sizes (12M–110M parameters) supports the conclusion that the effect is a property of the texts rather than an artifact of any particular model. DistilBERT is the only model showing a centroid increase for any pair (DE-CO), by 0.006, which is confined to the pair with the smallest overall effect and is negligible in magnitude.

**Table 7** Phase 3: intra-theory self-similarity (centroid).
SDI = 1.0 – centroid score. Theories ordered by SDI.

| Theory | Model | Centroid Score | SDI |
|---|---|---|---|
| *CO* | SBERT | 0.8781 | 0.1219 |
| | ALBERT | 0.8657 | 0.1343 |
| | DistilBERT | 0.8506 | 0.1494 |
| | RoBERTa | 0.8732 | 0.1268 |
| | TinyBERT | 0.8805 | 0.1195 |
| | MiniLM | 0.8724 | 0.1276 |
| | E5-base | 0.9847 | 0.0153 |
| | GTE-base | 0.9895 | 0.0105 |
| | **Average** | **0.8993** | **0.1007** |
| *VE* | SBERT | 0.8991 | 0.1009 |
| | ALBERT | 0.8782 | 0.1218 |
| | DistilBERT | 0.8380 | 0.1620 |
| | RoBERTa | 0.8837 | 0.1163 |
| | TinyBERT | 0.8902 | 0.1098 |
| | MiniLM | 0.8894 | 0.1106 |
| | E5-base | 0.9825 | 0.0175 |
| | GTE-base | 0.9869 | 0.0131 |
| | **Average** | **0.9060** | **0.0940** |
| *DE* | SBERT | 0.9489 | 0.0511 |
| | ALBERT | 0.9482 | 0.0518 |
| | DistilBERT | 0.9374 | 0.0626 |
| | RoBERTa | 0.9435 | 0.0565 |
| | TinyBERT | 0.9465 | 0.0535 |
| | MiniLM | 0.9371 | 0.0629 |
| | E5-base | 0.9912 | 0.0088 |
| | GTE-base | 0.9944 | 0.0056 |
| | **Average** | **0.9559** | **0.0441** |

We do not claim STS delivers a precise cardinal ground truth, nor does the STS evaluation literature itself, given known ceiling effects in human annotator agreement (Agirre et al., 2012; Cer et al., 2017). Accordingly, we hold ourselves to the standard the field already uses for its own gold-standard construction: directional and ordinal consistency, which is what the cross-model agreement (23/24), with the one principled exception) is offered as evidence for.

## 8.3 The semantic diffusion index

The average SDI ranking was CO > VE > DE; although the ordering of CO and VE varied across models, DE had the lowest SDI under all eight models. A higher SDI indicates that the raw document's semantic orientation was more strongly influenced by contrastive and meta-discursive material.

The divergence between volume of removal and SDI is instructive. Deontological ethics loses 64.2% of its raw word count but has an SDI (0.044) less than half of consequentialism's (0.101), which loses only 46.8%. Since DE's contrastive material is concentrated in a discrete section reviewing consequentialism, its directional pull is diluted by surrounding assertive sentences. CO's contrastive vocabulary is distributed through rule utilitarianism's incorporation of deontological reasoning, producing a larger cumulative centroid shift despite smaller volume. The SDI measures *semantic* impact on the document's orientation, not *lexical* volume.

The novelty of the SDI does not lie in the algebraic operation $(1 - \text{sim})$ itself, but in its use as an operational construct for quantifying semantic reorientation between a document's raw and high-level preprocessed representations. It converts the effect of preprocessing on a document's semantic orientation into a per-document diagnostic that is independent of any particular comparison document.

The SDI is method-agnostic at the conceptual level: it requires only a preprocessed version of the document and a similarity method, but places no constraints on the method itself. In the present study, it is operationalized as the additive complement of centroid cosine similarity across eight Transformer-based models. Cross-method validation using set-theoretic and vector-space methods remains a direction for future work.

### 8.4 Limitations and future work

The twelve rules are applied manually. This is labor-intensive but deliberate: the assertive–contrastive distinction requires subject-matter understanding. An automated approach—using large language models or argument mining methods—is a natural direction for future work. The twelve rules provide a specification against which automated classifiers can be validated, with the manual preprocessing serving as ground truth.

The underlying problem—semantic diffusion through discursive engagement—applies to any domain where texts discuss, compare, or argue against other positions: legal briefs, policy documents, academic articles, and editorial writing. The specific rules are necessarily domain- and purpose-dependent, because what counts as non-target content depends on the analytical objective. For example, Rule 4 removes cross-theory material when the objective is to compare substantive claims, although the same material would be informative in a study of discursive engagement; Rule 6 reflects the present study's objective of secular semantic comparison; and Rule 9 standardizes linguistic form where orthographic variation is treated as irrelevant, but would require reconsideration if such variation were itself analytically meaningful. The framework is therefore general, while its operational rules must be adapted to what the analysis is intended to preserve, remove, or normalize.

The general framework, however, is potentially domain-agnostic, and the SDI provides a corresponding per-document measure of the semantic reorientation induced by that preprocessing.

Pairwise similarity produces a single score that cannot distinguish genuine conceptual overlap from residual shared vocabulary. Preprocessing reduces the latter, but the resulting scores still merge all remaining sources of similarity into a single number. New analytical approaches are needed to address this limitation.

Per-rule ablation is not available because the rules emerged iteratively and no intermediate versions exist. The rules are also not independent: Rules 4 and 5 overlap substantially, while Rule 7 qualifies Rule 4. What the study offers instead is structural evidence: cross-phase consistency (SDIs are broadly consistent with pairwise changes), differential effect (volume and SDI diverge), and cross-model agreement (23/24 comparisons across eight architecturally diverse models). Per-rule sensitivity analysis remains a priority for future work.

## 9. Conclusion

This paper has identified a systematic gap in the text preprocessing literature: a lack of methods for isolating assertive content from discursive scaffolding prior to semantic similarity analysis. To address this gap, we proposed high-level text preprocessing: twelve rules, each with an explicit rationale, targeting specific sources of semantic diffusion and noise.

A three-phase experiment using eight Transformer-based STS models confirmed the central hypothesis. Preprocessing reduced centroid similarity across all three theory pairs, with 23 of 24 model–pair comparisons showing the expected decrease and cross-model agreement of 8-0, 8-0, and 7-1. The Semantic Diffusion Index (SDI) yielded an average ranking of consequentialism (0.101) > virtue ethics (0.094) > deontological ethics (0.044), although the ordering of CO and VE varied across models; DE had the lowest SDI across all eight models.

The analysis also confirmed that the volume of content removal and the semantic impact are not equivalent: deontological ethics lost more raw material than consequentialism did. However, it showed a substantially lower SDI, consistent with its contrastive content being more concentrated rather than distributed.

Three directions for future work are immediate. First, automation: the twelve rules provide a specification against which automated classifiers can be validated. Second, new analytical approaches are needed to distinguish genuine conceptual overlap from other sources of shared vocabulary that contribute to pairwise similarity scores. Third, cross-method SDI validation using STS methods from different families helps establish the operational extent of method agnosticism.

**Ethical Approval and Informed Consent by Author 1**:
This article does not contain any studies with human participants performed by any of the authors.
**Ethical Approval and Informed Consent by Author 2**:
This article does not contain any studies with human participants performed by any of the authors.

## Appendix A. Assessment of Alternative Philosophical Reference Sources

The raw corpus consists of three entries from the Stanford Encyclopedia of Philosophy (SEP): Virtue Ethics (Hursthouse & Pettigrove, 2023), Deontological Ethics (Alexander & Moore, 2021), and Consequentialism (Sinnott-Armstrong, 2023). A critical requirement is that each theory be represented by a consolidated, self-contained treatment of comparable length. SEP entries satisfy this: each exceeds 12,000 words (range: 12,801–13,259) within a single document. Alternative sources—including the Routledge Encyclopedia of Philosophy (Craig, 1998), the Encyclopedia of Ethics (Becker & Becker, 2001), Encyclopaedia Britannica, the Internet Encyclopedia of Philosophy, the Oxford Handbook of Ethical Theory (Copp, 2006), and the Bloomsbury Handbook of Ethics (Miller, 2023)—were evaluated but found unsuitable.

A description of the alternative corpora is as follows:

**Routledge Encyclopedia of Philosophy** distributes theoretical content across multiple shorter entries by subtopic and historical figure. Reconstructing a consolidated description would require aggregating content from multiple entries by different authors.

**Encyclopedia of Ethics** contains standalone entries but with radical length inequality: consequentialism with about 1,600 words, deontology with about 4,600, and virtue ethics with about 4,700. The 3:1 ratio would introduce severe structural asymmetry.

**Encyclopaedia Britannica** presents consequentialism and deontological ethics as two-paragraph glosses. Extended treatment appears under a general “Ethics” article, where theories are interwoven.

**The Internet Encyclopedia of Philosophy** lacks a standalone entry on deontological ethics. Consequentialism is fragmented across entries on utilitarianism and individual philosophers.

**Oxford Handbook of Ethical Theory** covers all three theories, each in a separate chapter, but with substantial length inequality (about 19,500 versus 10,500 words). Chapters are argumentative rather than expository.

**Bloomsbury Handbook of Ethics** lacks a standalone entry on deontological ethics. The topic is covered under “Kantian Ethics.” The consequentialism chapter adopts a heavily formalized, decision-theoretic style rather than an expository one.

Table A1 summarizes the assessment of each potential source across four dimensions. The *Consolidated* column indicates whether each theory is treated in a single, self-contained document. The *Comparable Length* column indicates whether the entries for the three theories are approximately equal in word count. The *Complete Coverage* column indicates whether all three theories are represented. The *Expository Style* column indicates whether the entries are written as neutral surveys. The *Suitable* column provides an overall judgment of each source's suitability for inclusion in this study.

**Table A1** Assessment of philosophical reference sources.
Only SEP satisfies all four criteria.

| Source | Consolidated | Comparable Length | Complete Coverage | Expository Style | Suitable |
|---|---|---|---|---|---|
| *SEP* | ✓ | ✓ | ✓ | ✓ | ✓ |
| *Routledge Encyclopedia of Philosophy* | ✗ | n.a.* | ✗ | ✓ | ✗ |
| *Encyclopedia of Ethics* | ✓ | ✗ | ✓ | ✓ | ✗ |
| *Encyclopaedia Britannica* | ✗ | ✗ | ✗ | ✓ | ✗ |
| *Internet Encyclopedia of Philosophy* | Partial | n.a. | ✗ | ✓ | ✗ |
| *Oxford Handbook of Ethical Theory* | ✓ | ✗ | ✓ | ✗ | ✗ |
| *Bloomsbury Handbook of Ethics* | ✓ | ~ | ✗ | ✗ | ✗ |

n.a.* not applicable

## Appendix B. Mathematical Formulation of Centroid Cosine Similarity and the Semantic Diffusion Index

### B.1 Notation

The following notation is used throughout this appendix.

$D$: a document represented as the ordered set of its sentences.
$n$: the number of sentences in $D$.
$s_i$: the $i$-th sentence of $D$, where $i = 1,2,\dots,n$.
$f$: a Transformer-based sentence encoding function.
$d$: the dimensionality of the embedding space.
$\boldsymbol{e}_i$: the embedding vector of sentence $s_i$, where $\boldsymbol{e}_i \in \mathbb{R}^d$.
$E_D$: the set of sentence embeddings for the document $D$.
$\boldsymbol{c}_D$: the centroid vector of the document $D$.
$D^{\mathrm{r}}$: the raw (unpreprocessed) version of the document $D$.
$E_D^{\mathrm{r}}$: the sentence embeddings for the document $D^{\mathrm{r}}$.
$D^{\mathrm{p}}$: the preprocessed version of the document $D$.
$E_D^{\mathrm{p}}$: the sentence embeddings for the document $D^{\mathrm{p}}$.
$n^{\mathrm{r}}, n^{\mathrm{p}}$: the number of sentences in $D^{\mathrm{r}}$ and $D^{\mathrm{p}}$, respectively.
$\mathrm{sim}(\boldsymbol{e}_i, \boldsymbol{e}_j)$: cosine similarity between two embedding vectors, $\boldsymbol{e}_i$ and $\boldsymbol{e}_j$.
$\mathrm{SDI}(D)$: the semantic diffusion index of the document $D$.

### B.2 Sentence-level encoding

A document $D$ is an ordered set of $n$ sentences:

$$D = \{s_1, s_2, \dots, s_n\}$$

A Transformer-based encoding function f maps each sentence to a point in a $d$-dimensional real-valued vector space:

$$f: s_i \to \boldsymbol{e}_i |\ \boldsymbol{e}_i \in \mathbb{R}^d$$

where

$$\boldsymbol{e}_i = (e_{i1}, e_{i2}, \dots, e_{id}) | e_{ij} \in \mathbb{R} \text{ for } j = 1,2,\dots,d$$

The set of all sentence embeddings for the document $D$ is denoted $E_D$:

$$E_D = \{\boldsymbol{e}_1, \boldsymbol{e}_2, \dots, \boldsymbol{e}_n\}$$

### B.3 Document centroid

The centroid $\boldsymbol{c}_D$ of the document $D$ is the arithmetic mean of its sentence embeddings:

$$\boldsymbol{c}_D = \frac{1}{n}\sum_{i=1}^{n} \boldsymbol{e}_i \mid \boldsymbol{e}_i \in \mathbb{R}^d$$

Component-wise, the $j$-th element of the centroid is:

$$c_{Dj} = \frac{1}{n}\sum_{i=1}^{n} e_{ij} \text{ for } j = 1,2,\dots,d$$

The centroid is the center of mass of the document's sentence embeddings in $\mathbb{R}^d$. It represents the document's overall semantic orientation: the single point in the embedding space that best summarizes where the document's sentences collectively reside.

### B.4 Cosine similarity between document centroids

Given two documents $A$ and $B$ with centroids $\boldsymbol{c}_A$ and $\boldsymbol{c}_B$, the cosine similarity is

$$\text{sim}(\boldsymbol{c}_A, \boldsymbol{c}_B) = \frac{\boldsymbol{c}_A \cdot \boldsymbol{c}_B}{\|\boldsymbol{c}_A\|\|\boldsymbol{c}_B\|}$$

where the dot product is

$$\boldsymbol{c}_A \cdot \boldsymbol{c}_B = \sum_{j=1}^{d} c_{Aj} c_{Bj}$$

and the Euclidean norm of the centroid, $\boldsymbol{c}_D$ of the document $D$ is

$$\|\boldsymbol{c}_D\| = \sqrt{\sum_{j=1}^{d} {c_{Dj}}^2}$$

Cosine similarity measures the angle between two vectors, independent of their magnitudes. In the general case, it ranges from $-1$ to +1, where +1 indicates identical orientation, 0 indicates orthogonality, and $-1$ indicates opposite orientation. For Transformer-based sentence embeddings, cosine similarities are often concentrated in a relatively narrow range due to embedding-space anisotropy (Ethayarajh, 2019). In this study, all observed centroid cosine similarities were non-negative; accordingly, the empirical range was [0, 1]. Therefore, the centroid similarity of the documents $A$ and $B$, denoted as $\mathrm{S_C}(A,B)$, is given as:

$$\mathrm{S_C}(A,B) = \text{sim}(\boldsymbol{c}_A, \boldsymbol{c}_B)$$

Furthermore,

$$-1 \leq \mathrm{S_C}(A,B) \leq 1 \text{ in theory, but}$$
$$0 \leq \mathrm{S_C}(A,B) \leq 1 \text{ in practice.}$$

### B.5 Effect of preprocessing on the centroid

Let $D^{\mathrm{r}}$ be the raw version of a document with $n^{\mathrm{r}}$sentences, and let $D^{\mathrm{p}}$ be its preprocessed version with $n^p$ sentences, where $n^{\mathrm{p}} < n^{\mathrm{r}}$. High-level preprocessing primarily removes sentences and passages, while Rules 8–11 may also modify retained text or introduce limited definitional content. Therefore, the preprocessed document is treated as a transformation of the raw document:

$$D^{\mathrm{p}} = \mathrm{T}(D^{\mathrm{r}})$$

The embedding sets of $D^{\mathrm{r}}$ and $D^{\mathrm{p}}$ becomes

$$E_D^{\mathrm{r}} = \{f(s) \mid s \in D^{\mathrm{r}}\}, \text{ and}$$
$$E_D^{\mathrm{p}} = \{f(s) \mid s \in D^{\mathrm{p}}\}$$

respectively.

The raw centroid is:

$$\boldsymbol{c}_D^{\mathrm{r}} = \frac{1}{n^{\mathrm{r}}}\sum_{i=1}^{n^{\mathrm{r}}} \boldsymbol{e}_i \mid \boldsymbol{e}_i \in E_D^{\mathrm{r}}$$

The preprocessed centroid is:

$$\boldsymbol{c}_D^{\mathrm{p}} = \frac{1}{n^{\mathrm{p}}}\sum_{i=1}^{n^{\mathrm{p}}} \boldsymbol{e}_i \mid \boldsymbol{e}_i \in E_D^{\mathrm{p}}$$

Most of the transformation consists of removing contrastive and meta-discursive content. The remaining modifications standardize or clarify retained content. Consequently, the difference between $\boldsymbol{c}_D^r$ and $\boldsymbol{c}_D^p$reflects the combined effect of content removal and limited textual transformation. The embedding vectors of contrastive sentences point toward the semantic regions of the rival theories they discuss. Their presence in the raw centroid's summation pulls $\boldsymbol{c}_D^{\mathrm{r}}$ toward those regions. Preprocessing removes these vectors, causing the centroid to shift away from the rival theories' regions and toward the document's assertive core.

### B.6 The semantic diffusion index

The semantic diffusion index quantifies the magnitude of the centroid shift induced by preprocessing. It is defined as the additive complement of the cosine similarity between a document's raw and preprocessed centroids:

$$\mathrm{SDI}(D) = 1 - \mathrm{sim}\left(\boldsymbol{c}_D^{\mathrm{r}}, \boldsymbol{c}_D^{\mathrm{p}}\right)$$

Because cosine similarity ranges from −1 to +1, the SDI theoretically ranges from 0 to 2:

$$0 \leq \mathrm{SDI}(D) \leq 2$$

In the present study, all raw–preprocessed centroid similarities are non-negative, so the observed SDI values lie between 0 and 1:

$$0 \leq \mathrm{SDI}(D) \leq 1$$

An SDI of 0 indicates that preprocessing did not alter the document's semantic orientation: the raw and preprocessed centroids point in the same direction. An SDI approaching 1 indicates that preprocessing produced a substantial reorientation: the preprocessing transformation produced a substantial reorientation of the document's semantic representation.

### B.7 Preprocessing and inter-document similarity

Let $A$ and $B$ be two discursive documents that discuss each other's subject matter. The raw centroid of $A$ is pulled toward $B$'s semantic region by $A$'s contrastive sentences about $B$, and vice versa. This mutual attraction inflates the similarity score:

$$\mathrm{S_C}(A^{\mathrm{r}}, B^{\mathrm{r}}) = \mathrm{sim}(\boldsymbol{c}_A^{\mathrm{r}}, \boldsymbol{c}_B^{\mathrm{r}})$$

After preprocessing, the contrastive vectors are removed from both documents. The centroids retreat toward their respective assertive cores:

$$\mathrm{S_C}(A^{\mathrm{p}}, B^{\mathrm{p}}) = \mathrm{sim}\left(\boldsymbol{c}_A^{\mathrm{p}}, \boldsymbol{c}_B^{\mathrm{p}}\right)$$

The expected effect of preprocessing is:

$$\mathrm{S_C}(A^{\mathrm{p}}, B^{\mathrm{p}}) \leq \mathrm{S_C}(A^{\mathrm{r}}, B^{\mathrm{r}})$$

The difference between the two scores quantifies the net change in inter-document similarity associated with high-level preprocessing:

$$\Delta(A, B) = \mathrm{S_C}(A^{\mathrm{r}}, B^{\mathrm{r}}) - \mathrm{S_C}(A^{\mathrm{p}}, B^{\mathrm{p}})$$

Under the semantic-diffusion hypothesis, the expected direction is $\Delta \geq 0$. A positive $\Delta$ indicates that the raw texts appear more similar than their preprocessed representations. Given that the a posteriori transformation audit shows that preprocessing was overwhelmingly subtractive, this pattern is consistent with the removal of discursive and other non-target material that had contributed to the raw similarity.

### B.8 The pairwise similarity matrix

Given two documents $A$ and $B$ with $m$ and $n$ sentences respectively, and corresponding embedding sets $E_A = \{\boldsymbol{e}_{A1}, \boldsymbol{e}_{A2}, \dots, \boldsymbol{e}_{Am}\}$ and $E_B = \{\boldsymbol{e}_{B1}, \boldsymbol{e}_{B2}, \dots, \boldsymbol{e}_{Bn}\}$, the pairwise similarity matrix $M$ is an $m \times n$ matrix whose entry at row $i$ and column $j$ is the cosine similarity between the $i$-th sentence of $A$ and the $j$-th sentence of $B$:

$$M_{ij} = sim\left(\boldsymbol{e}_{Ai}, \boldsymbol{e}_{Bj}\right) \text{ for } i = 1,2, \dots, m \text{ and } j = 1,2, \dots, n$$

Each entry $-1 \leq M_{ij} \leq 1$Mij measures the semantic similarity between one sentence from $A$ and one sentence from $B$.

### B.9 Mean pairwise sentence similarity

The mean pairwise sentence similarity $\mathrm{S_P}$ between documents $A$ and $B$ is the arithmetic mean of all entries in $M$:

$$\mathrm{S_P}(A, B) = \frac{1}{mn} \sum_{i=1}^{m} \sum_{j=1}^{n} M_{ij}$$

This metric treats every sentence pair as an equally weighted observation, where the total number of observations is $|M| = mn$.

### B.10 The noise floor

In a long discursive document, sentences address diverse subtopics. When both documents contain hundreds of sentences, the vast majority of the $mn$ pairs are semantically unrelated. Let the entries of $M$ be partitioned into two categories, signal pairs $M^{\sigma}$ and noise pairs $M^{\nu}$ so that

$$M_{ij}^{\sigma} = \begin{cases} M_{ij}, \text{ signal pair} \\ 0, \text{ otherwise} \end{cases}, \text{ and}$$

$$M_{ij}^{\nu} = \begin{cases} M_{ij}, \text{ noise pair} \\ 0, \text{ otherwise} \end{cases}$$

where signal pairs are those in which both sentences address related subtopics, and noise pairs are those in which the two sentences are semantically unrelated. Let $k$ denote the number of signal pairs and $(mn - k)$ the number of noise pairs. We define the noise floor as the mean similarity across noise pairs:

$$\nu = \frac{1}{mn-k}\sum_{i=1}^{m}\sum_{j=1}^{n} M_{ij}^{\nu}$$

and the signal mean as

$$\sigma = \frac{1}{k}\sum_{i=1}^{m}\sum_{j=1}^{n} M_{ij}^{\sigma} > 0$$

Since $k \ll mn$, signal pairs constitute only a small fraction of all sentence pairs. Consequently, their contribution to the overall mean is strongly attenuated:

$$\mathrm{S_P}(A,B) = \frac{k}{mn}\sigma + \frac{mn-k}{mn}\nu$$

For the signal component to exceed the noise component,

$$\frac{k}{mn}\sigma > \frac{mn-k}{mn}\nu$$

which requires

$$\frac{\sigma}{\nu} > \frac{mn-k}{k}$$

Because $k \ll mn$, the ratio $\frac{(mn-k)}{k}$ is large. Thus, $\sigma$ must be substantially greater than $\nu$ for signal pairs to dominate the mean pairwise similarity. Otherwise, the large number of noise pairs exerts a substantial influence on $S(A,B)$.

### B.11 The concentration effect

Now consider the effect of preprocessing on $\mathrm{S_P}(A,B)$. Let document $A$ have $m^{\mathrm{r}}$ sentences before preprocessing and $m^{\mathrm{p}}$ sentences after, where $m^{\mathrm{p}} < m^{\mathrm{r}}$. Similarly for document $B$, $n^{\mathrm{p}} < n^{\mathrm{r}}$. The matrix dimensions change from $m^{\mathrm{r}} \times n^{\mathrm{r}}$ to $m^{\mathrm{p}} \times n^{\mathrm{p}}$. Preprocessing removes two categories of sentences, each with a different effect on the matrix.

**Removal of meta-discursive sentences.** Meta-discursive sentences are about the document itself, not about its subject matter. They form noise pairs with virtually every sentence in the other document. Their removal eliminates rows or columns from $M$ that consist almost entirely of relatively low-similarity or baseline entries. This reduces the denominator $mn$ without proportionally reducing the number of signal pairs *k*. As a result, the ratio, $k\ /\ mn$, increases, and:

$$\mathrm{S_P}(A^{\mathrm{p}},B^{\mathrm{p}}) > \mathrm{S_P}(A^{\mathrm{r}},B^{\mathrm{r}})$$

This phenomenon is called the *concentration effect*. The mean pairwise score rises after preprocessing, not because the documents have become more semantically similar, but because the noise floor has been disproportionately reduced.

**Removal of contrastive sentences.** Contrastive sentences are more likely to form signal pairs with relevant sentences in the other document; their removal, therefore, tends to reduce the signal component as well as the pair-count denominator, $\sum M^{\sigma}$, and $mn$respectively. This, in turn, decreases the mean.

The net direction of change depends on which effect dominates. The concentration effect dominates when

$$\Delta(A,B) = \mathrm{S_P}(A^{\mathrm{r}},B^{\mathrm{r}}) - \mathrm{S_P}(A^{\mathrm{p}},B^{\mathrm{p}}) < 0$$

and signal reduction dominates when

$$\Delta(A,B) = \mathrm{S_P}(A^{\mathrm{r}},B^{\mathrm{r}}) - \mathrm{S_P}(A^{\mathrm{p}},B^{\mathrm{p}}) \geq 0$$

If the concentration effect from meta-discursive removal exceeds the signal-reduction effect from contrastive removal, then $\Delta < 0$: the score increases after preprocessing. This is the reason mean pairwise similarity is unsuitable for measuring the semantic impact of content-level preprocessing on long discursive documents.

### B.12 Why the centroid avoids the pair-count concentration effect

The centroid metric does not suffer from this problem because it does not compute pairwise sentence similarities. Instead, it reduces each document to a single vector, computed as the mean of its sentence embeddings, before computing similarity. The division by $n$ normalizes for document length. Removing a meta-discursive sentence whose embedding is near the centroid barely shifts $\boldsymbol{c}_A$. Removing a contrastive sentence $s_{Ac}$ whose embedding points toward $B$'s semantic region shifts $\boldsymbol{c}_A$ away from $B$'s region. The centroid is sensitive to the direction and amount of removed content, but unlike mean pairwise similarity, it is not directly affected by the $mn$ pair-count denominator.

Formally, the centroid after removing the sentence, $s_{Ac}$, from the document $A$ is:

$$\boldsymbol{c}'_A = \frac{1}{m-1}\left(\sum_{i}^{m} \boldsymbol{e}_{Ai} - \boldsymbol{e}_{Ac}\right)$$

The shift, $\boldsymbol{c}'_A - \boldsymbol{c}_A$, depends on the direction of $\boldsymbol{e}_{Ac}$ relative to $\boldsymbol{c}_A$. If $\boldsymbol{e}_{Ac}$ points toward $B$'s region (contrastive content), the centroid moves away from $B$. If $\boldsymbol{e}_{Ac}$ is near the centroid (noise content), the centroid barely moves.

This avoids the pair-count concentration mechanism while retaining sensitivity to genuine changes in document composition and semantic orientation.